# Intelligent Gradient Boosting Algorithms for Estimating Strength of Modified Subgrade Soil

Ismail B. Mustapha[1], Muyideen Abdulkareem[2]*, Shafaatunnur Hasan[1], Abideen Ganiyu[3], Hatem Nabus[1], Jin Chai Lee[2]

[1] Department of Computer Science, School of Computing, Universiti Teknologi Malaysia, Malaysia
[2] Department of Civil Engineering, Faculty of Engineering and Built Environment, UCSI University, Kuala Lumpur, Malaysia
[3] Department of Civil Engineering, British University of Bahrain, Saar, Bahrain

Email: *muyikareem@gmail.com

**Abstract**:
The performance of pavement under loading depends on the strength of the subgrade. However, experimental estimation of properties of pavement strengths such as California bearing ratio (CBR), unconfined compressive strength (UCS) and resistance value (R) are often tedious, time-consuming and costly, thereby inspiring a growing interest in machine learning based tools which are simple, cheap and fast alternatives. Thus, the potential application of two boosting techniques; categorical boosting (CatBoost) and extreme gradient boosting (XGBoost) and support vector regression (SVR), is similarly explored in this study for estimation of properties of subgrade soil modified with hydrated lime activated rice husk ash (HARSH). Using 121 experimental data samples of varying proportions of HARSH, plastic limit, liquid limit, plasticity index, clay activity, optimum moisture content, and maximum dry density as input for CBR, UCS and R estimation, four evaluation metrics namely coefficient of determination ($R^2$), root mean squared error (RMSE), mean absolute error (MAE) and mean absolute percentage error (MAPE) are used to evaluate the models' performance. The results indicate that XGBoost outperformed CatBoost and SVR in estimating these properties, yielding $R^2$ of 0.9994, 0.9995 and 0.9999 in estimating the CBR, UCS and R respectively. Also, SVR outperformed CatBoost in estimating the CBR and R with $R^2$ of 0.9997 respectively. On the other hand, CatBoost outperformed SVR in estimating the UCS with $R^2$ of 0.9994. Feature sensitivity analysis shows that the three machine learning techniques are unanimous that increasing HARSH proportion lead to values of the estimated properties respectively. A comparison with previous results also shows superiority of XGBoost in estimating subgrade properties.

**Notations**

| | |
|---|---|
| ANFIS | Adaptive neuro-fuzzy inference system |
| ANN | Artificial neural network |
| CaO | Calcium oxide |
| CBT | Catboost |
| CO2 | Carbo dioxide |
| DT | Decision tree |
| ELM | Extreme learning machine |
| FA | Fly ash |
| FFA | Firefly algorithm |
| GBDT | Gradient boosted decision trees |
| GBM | Gradient-tree boosting machine |
| GBR | Gradient boosting regressor |
| GEP | Gene expression programming |
| GGBS | Ground granulated blast furnace slag |
| GHG | Greenhouse gas |
| GRA | Grey relation analysis |
| HPC | High performance concrete |
| LightGBM | Light Gradient Boosting Machine |
| MAPE | Mean absolute percentage error |
| MARS | Multivariate adaptive regression splines |
| ML | Machine learning |
| MLP | Multilayer perceptron |
| MSE | Mean squared error |
| PC | Portland cement |
| R2 | Coefficient of determination |
| RBF | Radial basis function |
| RF | Random forests |
| RFG | Random forest regression |
| RMSE | Root mean squared error |
| RRMSE | Relative root mean squared error |

| SCBA | Sugarcane bagasse ash |
| SF | Silica fume |
| SVM | Support vector machine |
| SVR | Support vector regression |
| UHPC | Ultra-high performance concrete |
| XGBoost | Extreme gradient boosting |

## 1. Introduction

Road infrastructures are one of the most important civil structures that propels development of any nation. The performance/lifespan of pavements is attributed to the properties of the various layers that make up the pavement. A typical flexible pavement consists of four major layers that are surface, base, sub-base and subgrade. The movement of vehicles over the surface of the road pavement exerts different loads on each layer [1-3]. This is usually the compacted layer above the natural soil. The subgrade carries other layers above it and is therefore classified as the road foundation, thus, it is critical to adequately design to obtain the desired properties of the subgrade layer. These design properties include the California bearing ratio (CBR), unconfined compressive strength (UCS), and resistance value (R).

The CBR value is obtained by carrying out a CBR test that can be conducted in the field or in the laboratory [4]. The field CBR test involves penetration of a standard piston into the compacted soil sample, and the applied force is compared to the penetration depth. However, the field CBR test is costly as it involves the use of very expensive machines [5-7]. The laboratory CBR test on the other hand is time consuming and often leads to unreliable results due to sample disturbance, poor facilities and negligence during testing [8]. The UCS is the basic strength framework of soil that controls the deformation behaviour by using the load bearing capacity [9, 10]. In the laboratory determination of UCS of soil, it is difficult to obtain an undisturbed sample for a UCS test [11]. In addition, it is tedious, time consuming and expensive [12, 13]. The resistance value (R) expresses the resistance of the subgrade soil deformation due to vertical vehicular load. The deformation is given as a ratio of the transmitted lateral stress to the vertically applied stress. With values ranging from 0 to 100, it provides details about the thickness of the subgrade layer needed. Similar to the CBR, this test was invented by Carmany and Hveem, and test is carried out in the laboratory using a Hveem stabilometer. It involves soaking of the soil for 24 hours and then expelled by a specific applied load. Since it is often impossible to obtain a specimen to expel water at the specific load, this process is mostly cumbersome and time consuming given that several soil samples with different moisture contents need to be tested [14].

Owing to the aforementioned drawbacks, researchers have proposed using different machine learning (ML) techniques to estimate design properties of subgrade soils. These AI and ML approaches are not only quick, but are accurate and less tedious. These algorithms have been recently explored for the estimation of design properties of subgrade soils by considering certain soil properties, and they include support vector machine (SVM), adaptive neuro-fuzzy inference system (ANFIS), artificial neural network (ANN), and gene expression programming (GEP) [15, 16]. For example, Erzin and Turkoz [17] estimated the CBR of Turkish Aegean soils by using multiple regression and ANN models. The data consisted of 61 samples with a training/testing ratio of 80/20. The ten inputs feature that were considered are water content, specific gravity, dry density, coefficients of curvature and uniformity, and proportions of calcite, quartz, corund, feldspar, and amorphous minerals. The ANN model predicted values of CBR closer to the experimental results than those obtained by the multiple regression model giving $R^2$ of 0.9783 by the former as against 0.9387 by the latter. Trong [8] predicted the CBR of soils collected along a road project in Vietnam by applying three (3) random subspace optimization-based. The 214 samples (soil samples) were grouped according to AASHTO 145. The soil properties that included contents of coarse sand, gravel, silt clay, fine sand and organic, particle size distribution, plasticity index (PI), plastic limit (PL), liquid limit (LL), maximum dry density (MDD) and optimum moisture content (OMC) served as the input features to the models. Of the three models, the random subspace-based extra tree (RSS-ET) model performed better than the other two models (random subspace-based REPT and reduced error pruning tree) with an $R^2$ value of 0.968. Othman and Abdelwahab [7] predicted the CBR of 77 soil samples in Egypt with ANN. The optimization of the study was conducted by using 240 different architectures and hyperparameters. The results showed that the best performing ANN model is made-up of two hidden layer and employed a linear activation function. The ANN estimated the CBR value with good accuracy, achieving values of $R^2$ of 0.945, RMSE of 2.5, and MAE of 1.93.

Similarly, several studies have also focused on estimating of UCS of soils via ML techniques. Gunaydin [12] applied ANN models and different regression ML models to predict the UCS of soil. The study applied a dataset of 85 samples, and 9 input features that include moisture content, porosity, dry unit weight, unit weight, saturated unit weight, specific gravity, saturated degree, void ratio, and permeability. The ANN model outperformed all the regression analysis models, obtaining an $R^2$ value of 0.97. Sharma and Singh [18] presented a regression method for estimating the UCS of a lime-activated soil. The input variables were lime proportion, curing period, LL, PL, PI, ultrasonic wave velocity, potential of hydrogen, OMC and MDD. The results obtained were largely reliable, with $R^2$ of 0.96, RMSE of 25.89, and MAPE of 16.59. Akan and Keskin [11] estimated the UCS of clay soils obtained at different location by applying ANFIS and multiple linear regression (MLR) analyses. Soil properties applied are consistency limits, liquidity index, void ratio and fine grain ratio. Their study showed that optimum size of training and testing dataset dictates the estimation accuracy, and that the accuracy of ANFIS increased with normalized dataset. Their results showed that ANFIS outperformed MLR, with an $R^2$ value of 0.91 compared to 0.76 by the latter. Recently, Onyelowe [10] predicted the UCS of a cement-modified unsaturated soil using GEP, MEP and MLR, with an $R^2$ of 0.9872 achieved by GEP outperforming the duo of MEP (0.9845) and MLR (0.9805).

Furthermore, Kaloop [19] presented three ML techniques - Particle Swarm Optimization-based Extreme Learning Machine (PSO-ELM), Particle Swarm Optimization-based ANN (PSO-ANN) and Kernel ELM (KELM) – to predict the modulus of resilient ($M_R$) value of base soils. By using 704 data samples and 5 input features, the study was carried out by using a 70:30 training-test data split ratio. The study indicated that PSO-ELM achieved the best accuracy with an $R^2$ of 0.96, followed by PSO-ANN, and then KELM. Heidarabadizadeh [20] estimated the $M_R$ of soil materials with a hybrid SVM. The authors incorporated a colliding bodies optimization algorithm with SVM (SVM-CBO). The dataset consisted of 190 subbase and 272 non-cohesive subgrade materials, 75% of which was used for training and 25% for testing. The results showed that the $R^2$ values obtained for subbase, subgrade, and subbase-subgrade materials were 0.9883, 0.9792, and 0.9715 respectively. It was observed that increasing the MDD and uniformity coefficient increases the $M_R$ value. Khan [21] utilised ANN and GEP to predict the $M_R$ value of subgrade materials. The study involved stabilizing the subgrade materials by using cementitious materials and CaO. The input parameters are calcium oxide ratio, number of wet–dry cycles, OMC, deviator stress and confining pressure. The study applied 70/30 approach for the training/testing dataset. The study showed that ANN outperformed GEP with a better test $R^2$ value of 0.94 compared to 0.76 by the latter. The sensitivity analysis indicated that the OMC was the most dominant property in estimating the $M_R$ value, followed by the calcium oxide ratio and the number of wet–dry cycles. Recently, Ahmad [22] presented a study that compared three ML methods for predicting the resistance value (R) of a subgrade soil. The ML algorithms explored are M5P, SVM and Gaussian process regression (GPR). Using 121 data samples, the results showed that GPR produced the best test results with an $R^2$ of 0.9996, marginally outperforming SVM (0.9955) and M5P (0.9939).

Other related studies have also reported the prediction of multiple soil mechanical properties. For example, Onyelowe et al. Onyelowe [23] applied three different ANN models in estimating the properties of an clay soil that was modified with hydrated-lime activated rice husk ash (HARHA). The properties of the soil predicted are CBR, UCS, and R. Using a total of 121 data samples of 7 input features each, three algorithms that include Levenberg–Muarquardt backpropagation ANN, conjugate gradient and Bayesian programming, were explored for the prediction of the soil properties. Experimental findings showed that the Bayesian Programming model outperformed other techniques with the needed iteration to minimize error, while the Levenberg–Muarquardt Backpropagation provided the most accurate results with the least iterations. Also, Onyelowe [24] presented a sequel to this study, where GEP and MLR were employed for improved performance.

The aforementioned studies show that several ML techniques have been applied to estimate design properties of soils. Nevertheless, the quest for improved accuracy has inspired using more advanced ML techniques that includes gradient boosting algorithms that enhance the efficiency of separate base learners by together [25-27]. For instance, Cao [28] predicted the UCS of rock tunnel in Malaysia using an XGBoost with the firefly algorithm. Forty-five rock samples from the site were collected for this study. The XGBoost-FA provided results with an $R^2$ of 0.989, outperforming SVM and RBFN models that achieved $R^2$ of 0.972 and 0.886 respectively. Also, Eyo [29] applied XGBoost to estimate the CBR of soil stabilized with cement. The algorithm performed satisfactorily with $R^2$ and MAPE values of 0.924 and 2.032 respectively.

To avoid the dilemma of time-consuming, tedious and inaccurate laboratory and field testing, multitude of studies reported above have shown the importance of using ML techniques to accurately estimate the mechanical properties of soils for construction purposes. However, despite the improved performance of gradient boosting algorithms over conventional techniques like ANN, SVM, MLR and GEP that have been reported [30, 31], few studies have studied the comparative performance of boosting algorithms for estimation of mechanical properties of subgrade soils with stabilizing additives. Hence, in this study, two boosting ML algorithms – CatBoost and Xtreme gradient boosting (XGBoost) – and support vector regression (SVR) model, are applied in predicting the design properties of subgrade soil stabilized with hydrated-lime activated rice husk ash-treated (HARSH). The mechanical properties estimated (outputs) are the CBR, UCS and R, by using three (3) ML models for each estimation task. A total of 121 samples with 7 inputs that include HARSH proportion, LL, PL, PI, OMC, clay activity (AC) and MDD parameters are used in developing the models. Evaluation of the performance of each model is done using 4 popular statistical measures to have a broad perspective of the performance of each method (RMSE, MAE, MAPE and $R^2$). Likewise, a sensitivity study is undertaken in order to establish the important input parameters and their contribution, followed by a comparison with results obtained in past studies.

The study in the following key contributions:
- Prediction of the CBR, UCS and R of subgrade soil modified with hydrated-lime activated rice husk ash-treated (HARSH) using CBT and XGBoost.
- A comparison of results obtained from gradient boosting algorithms (CBT and XGB) for the prediction of the CBR, UCS and R of hydrated-lime activated rice husk ash-treated (HARSH) subgrade soil.
- An intuitive insight into the importance and contribution of input features for the estimation of the mechanical properties of HARSH subgrade soil.
- Comparison of results obtained in past studies.

## 2. Computational Methods
### 2.1 Support Vector Regression

Support vector regression (SVR) is a regression adaptation of support vector machine that applies ε-insensitive loss function for the penalization of data when higher than ε [32]. For any data $\{(x_m, y_m)\}_{m=1}^{n}$, where $x_m \in \mathbb{R}^d$ and $y_m \in \mathbb{R}$ are respectively the $m-th$ input and the corresponding real-valued output. $n$ is the sample size. The output is given as:

$$g(x_m) = \omega^T \varphi(x_m) + b \quad (1)$$

$\varphi$ is nonlinear mapper of input $x_m$, $w$ stands for the weight coefficient and $b$ is bias term. Flatness of (1) is ensured by minimizing the norm of $w$ as in (2).

$$min \frac{1}{2}\|\omega\|^2,$$
$$s.t \begin{cases} y_m - w^T \varphi(x_m) - b \leq \varepsilon, \\ y_m - w^T \varphi(x_m) - b \geq \varepsilon, \end{cases} \quad (2)$$

Introducing the two slack variables ($\xi$ and $\xi^*$) that are the upper and lower deviations to penalise the ε-insensitive band give (3)

$$min \frac{1}{2}\|\omega\|^2 + C \sum_{m=1}^{n}(\xi_m + \xi_m^*),$$
$$s.t \begin{cases} y_m - w^T \varphi(x_m) - b \leq \varepsilon + \xi_m, \\ y_m - w^T \varphi(x_m) - b \geq \varepsilon - \xi_m^*, \\ \xi_m, \xi_m^* \geq 0, m = 1,\ldots, n, \end{cases} \quad (3)$$

$C$ is the penalty parameter indicating the trade-off between the empirical risk and the regularization term. The generic equation is by Equation. (4) as:

$$g(x) = \sum_{m=1}^{n}(\beta_m - \beta_m^*) K(x_m, x) + b \quad (4)$$

$\beta_m$ and $\beta_m^*$, and $K(x_m, x)$ are the non-zero Lagrange multipliers and kernel function respectively. A range of kernel functions are tested and the Radial Basis Function kernel (RBF) to yield the best performance in this study.

$$K(x_m, x_i) = exp(-\gamma ||x_m - x_i||^2), \qquad (5)$$

$\gamma$ is the width parameter of RBF.

## 2.2 XGBoost

Xtreme Gradient Boosting (XGBoost) is an optimized variant of gradient boosting that combines the predictions of several "weak" classification and regression tree learners to develop a "strong" learner using additive training strategies [33]. It is known to prevent overfitting effectively via simple function that merges losses and regularization terms. The regularized optimization objective can be given as:

$$Obj = \sum_m^n l(y_m, \hat{y}_m) + \sum_k^K \Omega(f_k) \qquad (6)$$

$l$ is loss function that measures the difference between the estimated $\hat{y}_m$ output and the experimental, $y_m$,; $\Omega$ is regularization term which is defined as:

$$\Omega(f) = \gamma T + \frac{1}{2}\lambda \sum_{i=1}^{T} w^2 \qquad (7)$$

$T$ and $w$ are the leaves and the score on each leaf respectively; $\gamma$ and $\lambda$ are constants of the degree of regularization.

## 2.3 CatBoost

CatBoost (CBT) is an ML that leverages gradient boosting on decision trees. It is a distinctive gradient boosting decision tree implementation with special ability of handling features categorically [34]. Its advanced algorithmic introduced in CBT are an innovative algorithm for processing categorical features, and implementing ordered boosting.

## 3. Dataset

The dataset used in this study are from [23]. The data is made up of 121 samples with seven input features namely, HARSH, LL, PL, PI, Clay activity (CA), OMC, and MDD. These features are to be used for the prediction of CBR (%), UCS and R-value respectively. Table 1 shows the basic statistics of the 121 data samples of the inputs and the respective outputs. The statistics shown are the mean, standard deviation, maximum value, minimum value, upper quartile, middle quartile (median), and lower quartile to show the suitability and consistency of the dataset applied in this study.

**Table 1 Summary Statistics of Dataset**

|  | HARSH | Liquid limit (LL) | Plastic limit (PL) | Plasticity index (PI) | Optimum moisture content (OMC) | Clay activity (CA) | Maximum dry density (MDD) | CBR (%) | UCS | R-value |
|---|---|---|---|---|---|---|---|---|---|---|
| count | 121 | 121 | 121 | 121 | 121 | 121 | 121 | 121 | 121 | 121 |
| mean | 6.00 | 48.00 | 17.17 | 30.82 | 18.02 | 1.35 | 1.69 | 24.00 | 172.87 | 20.50 |
| std | 3.51 | 11.54 | 2.41 | 9.15 | 0.77 | 0.40 | 0.24 | 11.74 | 31.66 | 4.48 |
| min | 0 | 27 | 12.8 | 14 | 16 | 0.6 | 1.25 | 8 | 125 | 11.7 |
| 25% | 3 | 37 | 15 | 22 | 17.7 | 1 | 1.46 | 13.8 | 143 | 17.3 |
| 50% | 6 | 49 | 17.7 | 31 | 18.2 | 1.4 | 1.69 | 22.8 | 172 | 20.9 |
| 75% | 9 | 59 | 19 | 40 | 18.55 | 1.71 | 1.95 | 34 | 195 | 24 |
| max | 12 | 66 | 21 | 45 | 19 | 2 | 1.99 | 44.6 | 232 | 27 |

## 4. Experimental Design

The setup experiment in this study is shown in Fig. 1. Data normalization is carried out to avoid numerical overflow and placing the input variables on hold within a uniform range. Care is taken when splitting the dataset into training and testing partitions before the dataset is normalised to prevent data leakage. The input variables are normalised with values ranging from 0 to 1. It is noteworthy that the training-test split ratio adopted in this research is 70:30.

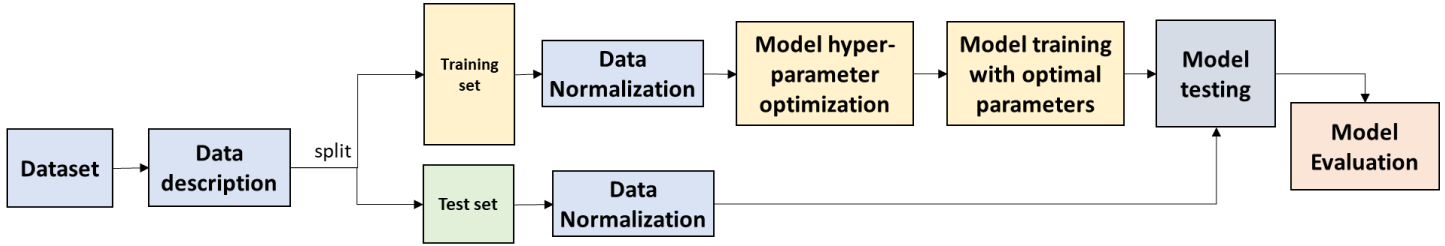

**Figure 1.** Experimental Setup

Another important stage in the experimental setup is the optimization of the hyperparameters of each ML model. A hyperparameter optimization is carried out for each model via training split of the dataset. Through this, the model's access to the test partition is restricted before the testing phase. Optimal performance of each model is ensured by using a wide range of tuneable hyperparameters. All available combination of values in a specific range of each hyperparameter is comprehensively searched to train each model by using 5-fold cross validation. This means that the training set is thereafter split to 5 equals with each applied to test the trained model performance while the other 4 use a combination of hyperparameters at a time. The optimal model hyperparameters is obtained via the hyperparameters combination with the lowest average mean squared error. Thus, applied to train the entire training data before testing with test partition set aside. The optimal hyperparameter for each model is presented in Table 2 for each prediction task.

**Table 2.** Optimal Hyperparameters for Models

|  | CBR (%) | UCS | R-value |
|---|---|---|---|
| **SVR** | SVR1 | SVR2 | SVR3 |
| $C$ | 52 | 500 | 75 |
| gamma | 0.99 | 0.96 | 0.9 |
| epsilon | 0.001 | 0.001 | 0.002 |
| kernel | rbf | rbf | rbf |
| **CatBoost** | **CBT1** | **CBT2** | **CBT3** |
| n_estimators | 289 | 500 | 493 |
| colsample_bytree | 0.4 | 1 | 1 |
| depth | 7 | 12 | 5 |
| subsample | default | default | 0.9 |
| l2_leaf_reg | 0.48 | 0.21 | 0.03 |
| learning_rate | 0.05 | 0.05 | 0.04 |
| **XGBoost** | **XGB1** | **XGB2** | **XGB3** |
| n_estimators | 349 | 359 | 242 |
| colsample_bytree | 0.5 | 0.7 | default |
| max_depth | 6 | 6 | 7 |
| subsample | 0.7 | 1 | 0.7 |
| reg_lambda | 1.1 | 1.21 | 0.06 |
| learning_rate | 0.03 | 0.04 | 0.03 |

## 5. Results and Discussion

The experimental results of the performance of each model on the three investigated estimation tasks (CBR, UCS, R-value) are presented in Table 3 in terms of four evaluation metrics with the best result being in bold for each metric. The results are discussed relative to each task in the following subsections.

### 5.1 Estimation of CBR (%)

The performances (training and testing) of the SVR1, CBT1 and XGB1 models in estimating CBR (%) is generally very impressive with the training outperforming their corresponding testing. In terms of $R^2$ which indicates the level of correlation between the estimated CBR (%) values and their respective experimental values. The training and testing of XGB1 (1.0 and 0.9999) marginally surpasses those of SVR1 (0.9998 and 0.9997) and CBT1 (1.0 and 0.9994) respectively. Figure 2 pictorially illustrates this using a scatter plot of the experimental against the predicted CBR (%) with a line of best fit (red line) for the training and testing phases. As can be observed from the plots, the line of best fit for each model passes through all the points in the plots reflecting how well these models generally perform. However, in terms of the test performances, the XGB1 model fits the data best as the points are more aligned on the line compared to SVR1 and CBT1.

**Table 3 Training and Test Performances of Models**

|  | Training | | | | Test | | | |
|---|---|---|---|---|---|---|---|---|
|  | ↑$R^2$ | ↓RMSE | ↓MAE | ↓MAPE | ↑$R^2$ | ↓RMSE | ↓MAE | ↓MAPE |
|  | CBR (%) | | | | | | | |
| SVR1 | 0.9998 | 0.1739 | 0.1125 | 0.0049 | 0.9997 | 0.2191 | 0.1605 | 0.0080 |
| CBT1 | **1.0000** | 0.0818 | 0.0652 | 0.0032 | 0.9994 | 0.2868 | 0.2135 | 0.0101 |
| XGB1 | **1.0000** | **0.0277** | **0.0197** | **0.0011** | **0.9999** | **0.1119** | **0.0926** | **0.0048** |
|  | UCS | | | | | | | |
| SVR2 | 0.9997 | 0.5111 | 0.2813 | 0.0016 | 0.9988 | 1.1039 | **0.5466** | **0.0031** |
| CBT2 | **1.0000** | **0.0150** | **0.0125** | **0.0001** | 0.9994 | 0.7985 | 0.6082 | 0.0035 |
| XGB2 | **1.0000** | 0.0988 | 0.0776 | 0.0005 | **0.9995** | **0.7343** | 0.6081 | 0.0039 |
|  | R-Value | | | | | | | |
| SVR3 | 0.9998 | 0.0593 | 0.0432 | 0.0023 | 0.9997 | 0.0794 | 0.0672 | 0.0036 |
| CBT3 | **1.0000** | 0.0186 | 0.0156 | 0.0008 | 0.9996 | 0.0995 | 0.0764 | 0.0043 |
| XGB3 | **1.0000** | **0.0097** | **0.0073** | **0.0004** | **0.9999** | **0.0556** | **0.0444** | **0.0023** |

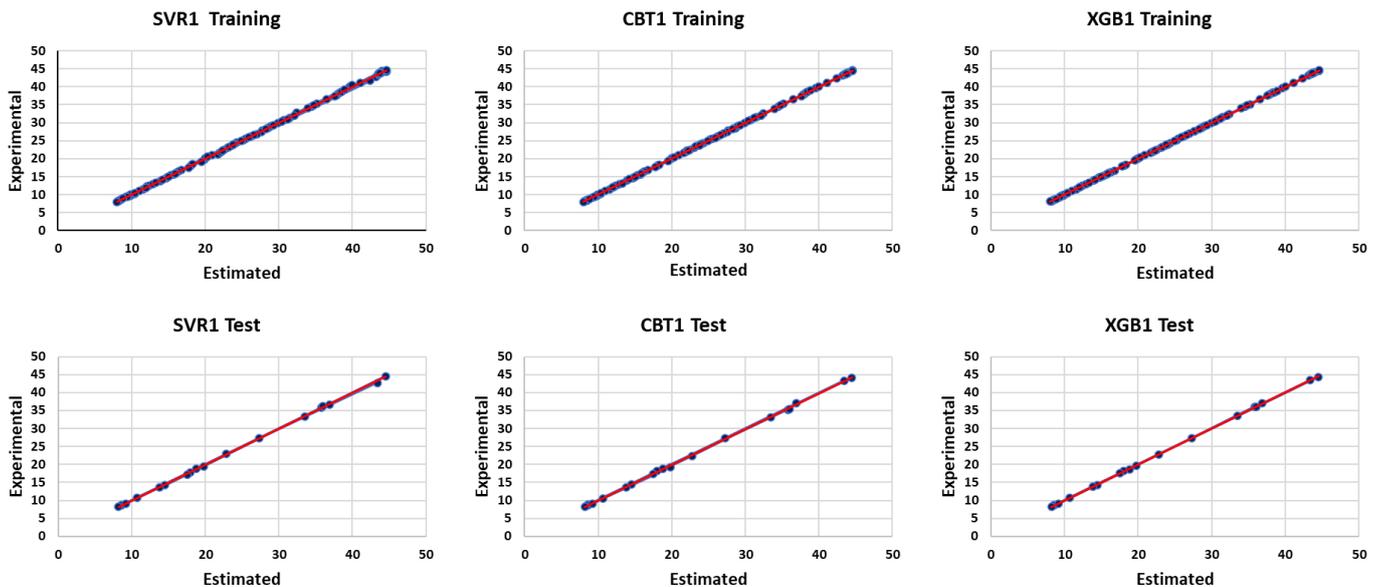

**Figure 2.** Training and Test Phases of CBR of the Experiment and Prediction

Similarly, the RMSE, MAE and MAPE measures that seek to quantify errors/differences between the experimental and estimated values, the training and testing performances of XGB1 (RMSE = 0.0277 and 0.1119; MAE = 0.0197 and 0.0926; MAPE = 0.0011 and 0.0048) are respectively better than those of SVR1 (RMSE = 0.1739 and 0.2191; MAE = 0.1125 and 0.1605; MAPE = 0.0049 and 0.0080) and CBT1 (RMSE = 0.0818 and 0.2868; MAE = 0.0652 and 0.2135; MAPE = 0.0032 and 0.0101). Although SVR1 produces a higher training error in terms of theses metrics compared to XGB1 and CBT1, it generalizes to the test data better that CBT1 given its lower test error but not XGB1. Figure 3 shows the superimposed error plots of each of these models (training and test phases). The errors in the training and testing phases of each model are estimated by the difference between the predicted and experimental CBR values. Hence, the model shows better performance when the error plot is near zero. As seen in Figure 3, XGB1 shows the least deviation in the training phases whereas SVR1 shows the highest deviation in the same phase. In the test phase, XGB1 similarly shows the least deviation, exceeding the |0.2| mark on only one occasion (sample index 2). Although,

SVR1 produced the largest single deviation of 0.6578 (sample index 15) from zero, CBT1 produced the highest number of deviations from the zero mark by exceeding the |0.2| mark on eight occasions (samples indices 2, 3, 4, 9, 10, 12, 14 and 19) compared to four in SVR1 (samples indices 5,12,15,18).

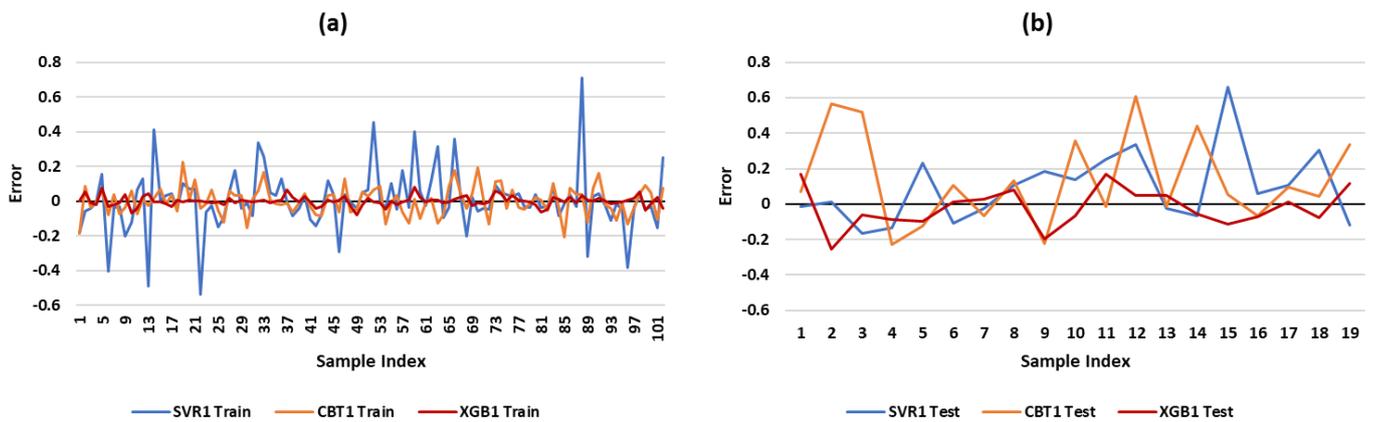

**Figure 3.** Error Plots for CBR (%) (a) Training (b) Test

## 5.2 Estimation of UCS

The training and testing performances of the SVR2, CBT2 and XGB2 models in UCS estimation are respectively very impressive with the respective training outperforming their corresponding test performance. In terms of $R^2$, all the models produced a near perfect estimation with the training and test performances of XGB2 (1.0 and 0.9995) negligibly surpassing those of SVR2 (0.9997 and 0.9988) and CBT2 (1.0 and 0.9994) respectively. Figure 4 shows the scatter plots of the experimental versus the predicted UCS with the corresponding line of best fit (red line) of both training and testing phases. The outstanding performance of each model is obvious in the fact that each point in the training and test scatter plots is diagonally aligned along the line of best fit. Nevertheless, in terms of the test performances which reflects the true model performance, the points are more aligned on the XGB2 model compared to SVR2 and CBT2; showing that XGB2 fits the data best, followed by CBT2 then SVR2.

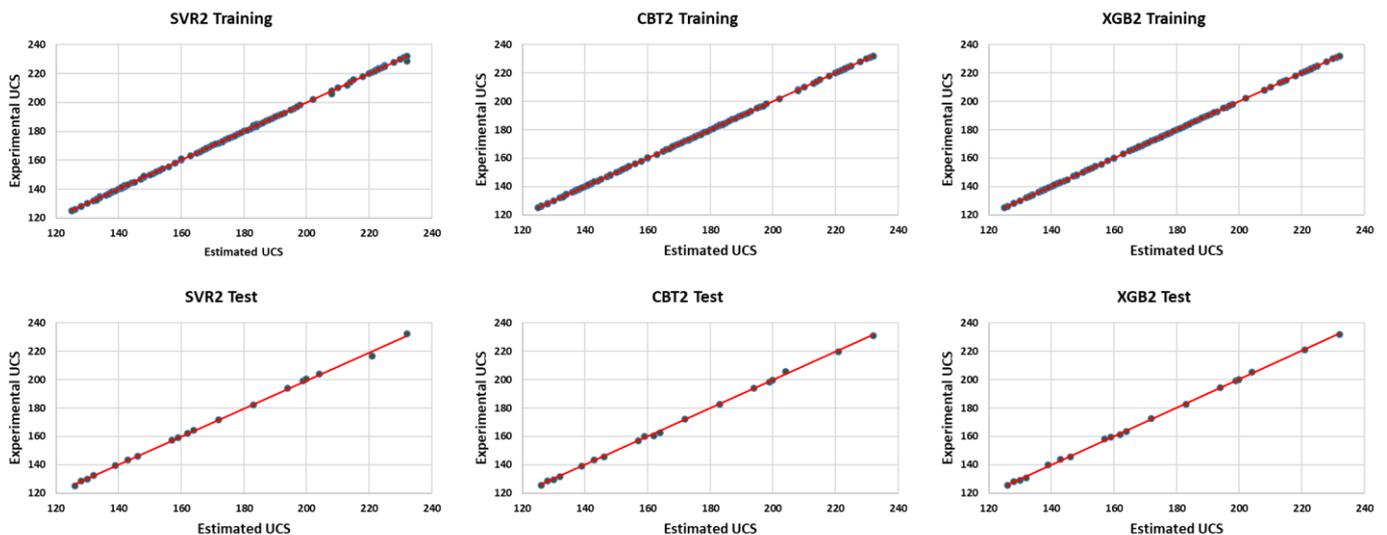

**Figure 4.** Training and Test Phases of UCS of the Experiment and Prediction

Regardless of the notable results provided by the models, there still exist minimal estimation errors that are can be captured by error measures like RMSE, MAE and MAPE. In terms of RMSE, XGB2 (RMSE = 0.0988 and 0.7343) produced a better training and test performance than SVR2 (RMSE = 0.5111 and 1.1039) and CBT2 (RMSE = 0.0150 and 0.7985). However, in terms of MAE and MAPE, despite producing the highest training error, SVR2 (MAE = 0.2813 and 0.5466; MAPE = 0.0016 and 0.0031) produced lower test errors than XGB2 (MAE = 0.0776 and 0.6081; MAPE= 0.0005 and 0.0039) and CBT2 (MAE = 0.0125 and 0.6082; MAPE = 0.0001 and 0.0035). This is further illustrated in Figure 5 where the superimposed error plots of the training and testing phases shown. The training error plot as in Figure 5(a) shows SVR2 (yellow line) to be the least performant given obvious higher number of deviations from the zero-mark compared to CBT2 and XGB2 models. However, in terms test performance as in Figure 5(b), while SVR2 produced the highest single deviation of 4.5065 from the zero mark on samples index 14, its errors on the other sample indices are below |1| with the highest being 0.7907. This is in contrast with XGB2 and CBR2 that both exceed |0.9| five (samples indices 1, 4, 5, 8 and 18) and six (sample indices 1, 2, 5, 9, 12 and 15) times respectively.

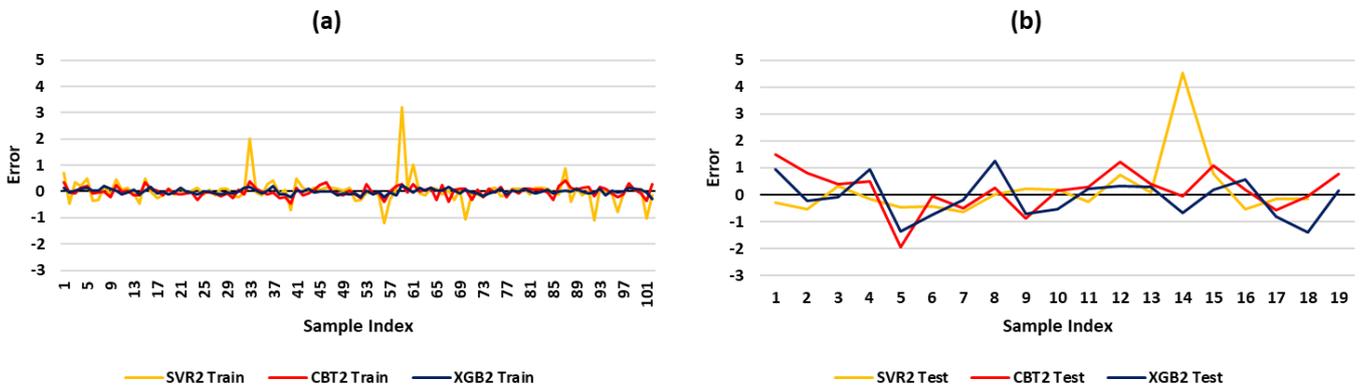

**Figure 5.** Error Plots for UCS (a) Training (b) Test

## 5.3 Estimation of R-Value

In estimating the R-value, the SVR3, CBT3 and XGB3 models performed impressively in both training and test phases with the respective training outperforming the corresponding testing. The performance of the models in terms of $R^2$ was near perfect as scores of 1.0 and 0.9999, 1.0 and 0.9996, and 0.9998 and 0.9996 for the XGB3, CBT3 and SVR3 models were obtained for both training and testing. Figure 6 shows the scatter plots of the experimental versus the estimated R-values with the corresponding line of best fit (red line) for the training and test phases of each model. The outstanding performance of each model is obvious in the fact that the line of best fit intersects each point in the training and test scatter plots of each model. A ranking of the of the models based on their test performance in terms of $R^2$ shows that XGB3 ranks first, followed by SVR3 then CBT3.

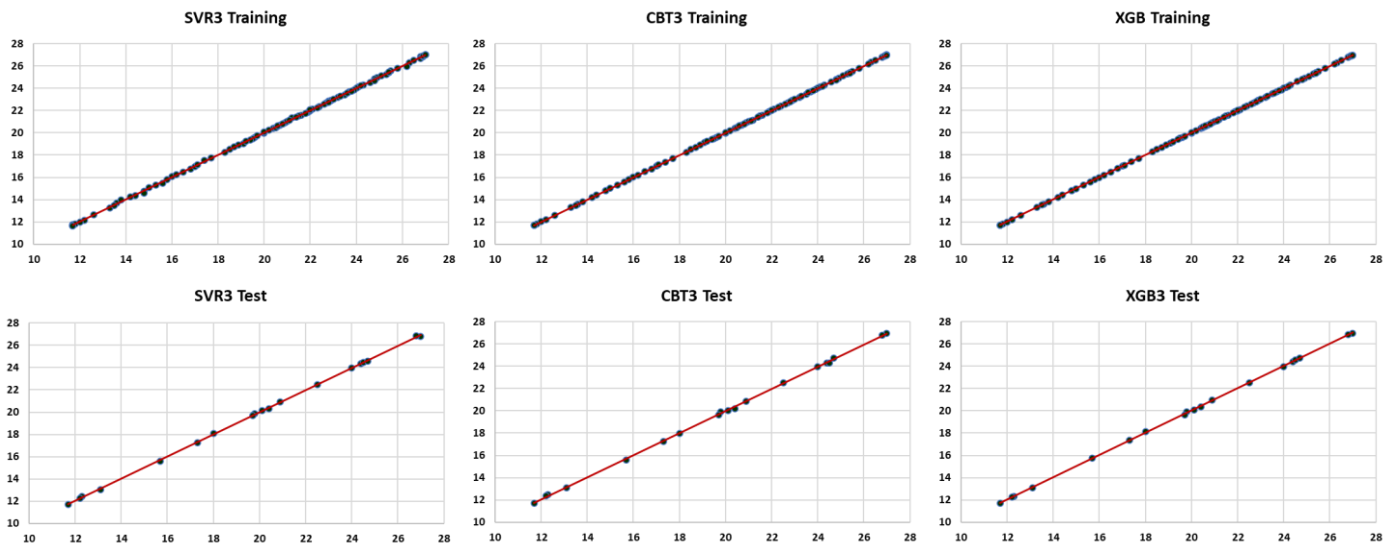

**Figure 6.** Training and Test Phases of R-value of the Experiment and Prediction

Also, the error measures (RMSE, MAE and MAPE) of the performance of each model on R-value estimation shows that XGB3 (RMSE = 0.0097 and 0.0556; MAE = 0.0073 and 0.0444; MAPE = 0.0004 and 0.0023) has lower training and test errors compared to SVR3 (RMSE = 0.0593 and 0.0794; MAE = 0.0432 and 0.0672; MAPE = 0.0023 and 0.0036) and CBT3 (RMSE = 0.0186 and 0.0995; MAE = 0.0156 and 0.0764; MAPE = 0.0008 and 0.0043) where a performance improvement range of 41% to 46% and 7% to 36% was recorded over CBT3 and SVR3 respectively in the test phase. The superior performance of XGB3 over SVR3 and CBT3 in the training and test phases is further illustrated using the error plots is Figure 7. In particular, the test performance in Figure(b) shows that While XGB3 shows highest deviation of -0.1207 and exceeds |0.1| on two occasions (sample indices 9 and 11), SVR3 and CBT3 produced the highest deviation of 0.1836 and 0.2073 respectively exceeding |0.1| on 5 instances (sample indices 4,5,6,12 and 15 by SVR3 and sample indices 3, 4, 6, 7 and 12 by CBT3).

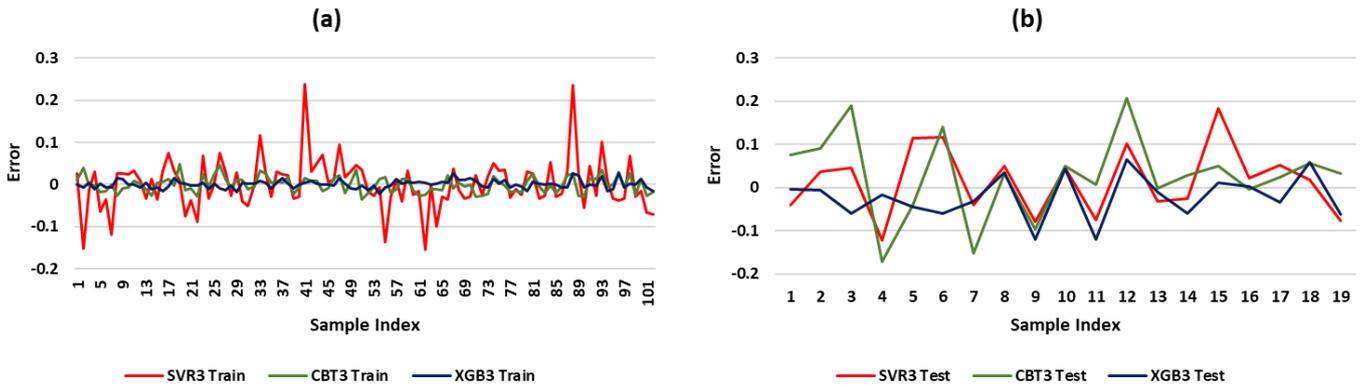

**Figure 7.** Error Plots for R (a) Training (b) Test

## 6. Average Model Performance

Multiple repetitions of machine learning experiments using different training and test splits can produce models that capture intrinsic variability in dataset and yield a more objective estimate of the performance and generalization of a model on a given task (). Thus, the experiments for the three estimation tasks (CBR, UCS and R-value) in this study were further repeated 10 times while varying the training and test splits each time. The average performance based on each evaluation metric for each model and estimation task is presented in Figure 8. For CBR estimation, XGB1 produced the best average performance given the highest $R^2$ score of 0.9996±0.0004 and lower errors based on RMSE, MAE and MAPE compared to SVR1 and CBT1. For UCS and R-value estimations, the $R^2$ scores of 0.9992±0.0006 and 0.9996±0.0002 produced by the SVR2 and SVR3 models for the respective estimation tasks rank higher than the scores produced by CBT2 and XGB2 as well as CBT3 and XGB3 respectively. Similarly, the SVR2 and SVR3 models produced lower RMSE, MAE and MAPE scores compared to CBT2 and XGB2, and CBT3 and XGB3 in UCS and R-value estimations respectively.

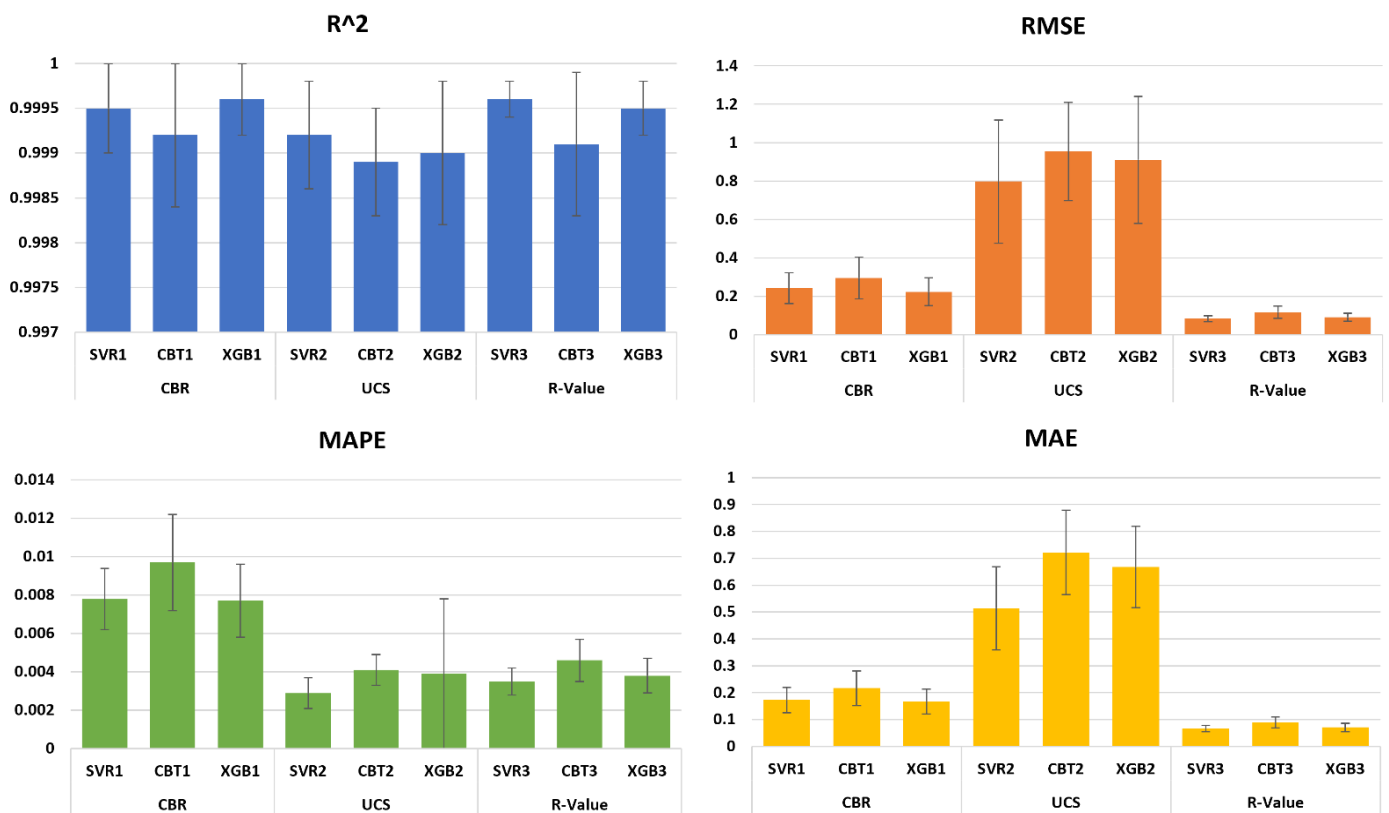

Figure 8. Average Results Across 10 Repetitions of Experiment

## 7. Sensitivity Study

A sensitivity study of each input feature to the output estimations of each model is presented here to understand how changes in input affects the target output. In this study, partial dependence plots (PDPs) [35] are used to analyse and visualize the relationship between target output and a specific input feature independent of all other input features for each estimation task. As shown in Figure 9, sensitivity analysis with PDPs makes it possible to see how changes in each input feature impact the model's estimation for each target variable while holding other features constant. The sensitivity analysis for each estimation task is presented in what follows. The range of each input feature and the output considered for the PDPs are as described in Table 1.

### 7.1 CBR Estimation

The PDPs for SVR1, CBT1 and XGB1 are respectively shown for CBR estimation across Figure 9(a) to (c). For SVR1, it can be observed in Figure 9(a) that HARSH proportion and maximum dry density (MDD) both exhibit a positive relationship with CBR, with the model predicting higher CBR as both quantities respectively increase. On the other hand, a negative or inverse relationship can be observed between the predicted CBR and the trio of PI, LL and PL, given increment in the values of each of these features resulting in corresponding lower CBR prediction by SVR1, though to a lesser degree with PL. As the OMC increases from 17.7 to 18, SVR1 predicts minimal increase in the CBR (22-24), however, further increase in OMC beyond this result in marginal reduction in the CBR predicted by the model. Likewise, as clay activity (AC) increases, SVR1 predicts minimal decrease in CBR from 25 to around 23.

For CBT1, Figure 9(b) shows that the PDPs of CBR vs HARSH proportion and CBR vs MDD respectively exhibits an upward trend, indicating an increase in predicted CBR as these two input features increase, hence a positive relationship. In contrast, a downward trend can be observed in the PDPs of CBR vs PI, CBR vs LL, CBR vs PL and CBR vs CA respectively, indicating that as these input features respectively increase, the predicted CBR reduces, hence an inverse relationship. In relation to the PDP of CBR vs OMC, increase in OMC from the minimum value of 16 to around 18.66 result in a relatively constant CBR value of about 23 predicted by CBT1, further OMC increment result in prediction of lower CBR values.

In relation to XGB1, the PDP of CBR vs HARSH proportion shows that increase in HARSH proportion result in the model predicting higher CBR. Similar trend can be observed in the PDP of CBR vs MDD, albeit increasing MDD only result in marginal increase in the predicted CBR. On the other hand, the PDPs of CBR vs PI and CBR vs LL respectively exhibit inverse relationship, with respective increase in PI and LL yielding corresponding lower CBR predictions by the model. While increasing values of PL and CA respectively appear to result in lower CBR predictions which suggests an inverse relationship, only a marginal reduction can be observed in the predicted CBR from 25. On the contrary, increasing value of OMC only produces a constant CBR prediction.

### 7.2 UCS Estimation

The PDPs for SVR2, CBT2 and XGB2 are respectively shown for UCS estimation from Figure 9(d) to Figure 9(f). Although, the range of the values of CBR and UCS are very disparate, the trend in the relationship between the respective input features and the predicted UCS values by the SVR2 model is similar to that of SVR1 model relative to CBR, with UCS vs HARSH and UCS vs MDD exhibiting a positive relationship whereas, the PDPs of UCS vs LL, UCS vs PL and UCS vs PI show an inverse relationship. In relation to the PDP of UCS vs OMC, a steady increment can be observed in the predicted UCS as OMC increases, however, further increase in OMC beyond 18 leads to steady decrease in predicted UCS values.

For CBT2, the trend displayed by the PDPs in Figure 9(e) is highly comparable to CBT1 (Figure 9(b)), as increasing values of HARSH proportion and MDD respectively result in an increase in predicted UCS values whereas, increasing values of PI, LL, PL and CA respectively results in the model predicting lower UCS values. Also, the PDP for UCS vs OMC shows that increase in OMC largely produces a constant value in predicted UCS until around OMC values of 18 – 18.6 where the predicted UCS witnesses a sharp increase, further OMC increment result in prediction of lower UCS values by CBT2.

The PDPs in Figure 9(f) shows that the XGB2 model is mostly responsive to changes in HARSH proportion and LL, with increase in HARSH mostly yielding prediction of high UCS whereas, increase in LL generally result in the prediction of lower UCS values by the same model. A slowly decreasing pattern can also be observed in predicted UCS as PI increases, but the UCS values appear to plateau between PI values of 40 - 45.

In contrast, the PDPs of UCS vs PL, UCS vs OMC, UCS vs CA and UCS vs MDD respectively show that the XGB2 model predicts the same UCS value in response to changes in PL, OMC, CA and MDD values.

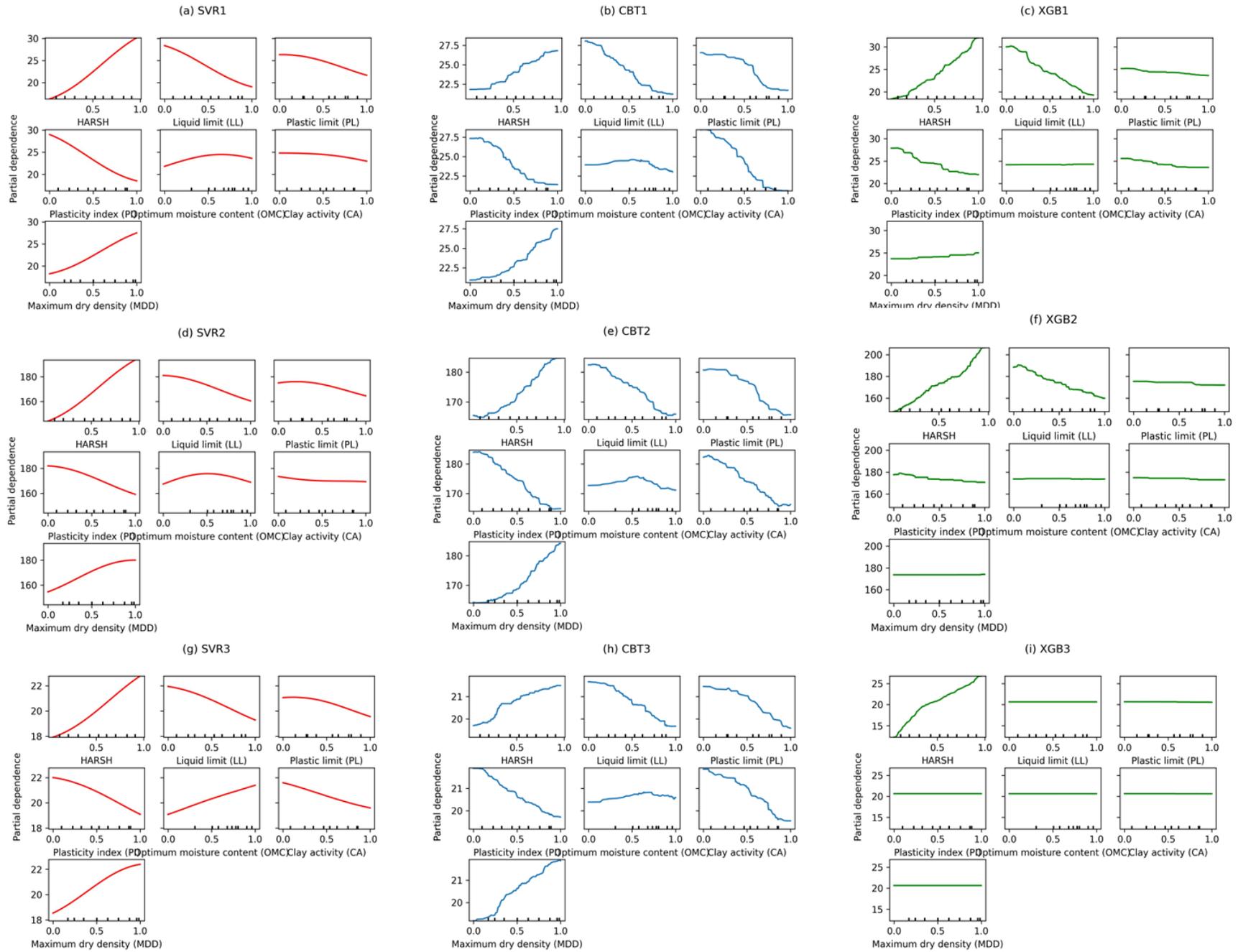

**Figure 9**. Sensitivity Analysis of Input Features to CBR (a-c) UCS (d-f) and R-value(g-i) Estimation Models

## 7.3 R-Value Estimation

Also, the PDPs for SVR3, CBT3 and XGB3 are respectively shown for R-value estimation from Figure 9(g) to Figure 9(i). For SVR3, a positive relationship can be observed in the PDPs of R-value vs HARSH, R-value vs MDD and R-value vs OMC, with increasing values of each input variable mostly resulting in prediction of higher R-values. On the contrary, the PDPs of R-value vs LL, R-value vs PL, R-value vs PI and R-value vs CA respectively show an inverse relationship, where increasing values of each input variable result in varying degree of lower R-values being predicted by the SVR3 model.

For CBT3, the trend displayed by the PDPs in Figure 9(h) is highly comparable to CBT1 and CBT2 (Figure 9(b) and Figure 9(e)), as increasing values of HARSH proportion and MDD respectively show a positive relationship with predicted R-values whereas, increasing values of PI, LL, PL and CA respectively exhibit inverse relationship with predicted R-values. In contrast, the PDP for R-value vs OMC shows that increase in OMC result in minimal but constant increase in predicted R-values until OMC value of 18.6, after which further OMC increment result in prediction of lower R-values by CBT3.

For XGB3, Figure 9(i) shows that only the PDP for R-value vs HARSH produces an insightful trend, with changes in HARSH proportions resulting in prediction of high R-values. In contrast, increment in the remaining inputs respectively result in the prediction of a constant R-value by the XGB3 model.

Overall, the trend shown by the PDPs across the prediction tasks for the SVR and CBT based models is mostly comparable. Although XGB-based models differ considerably, all the models are unanimous that increasing HARSH proportion results in prediction of higher values of CBR, UCS and R respectively. It should be noted that these reported findings are limited by the range of values of the input and output features of the dataset, coupled with the assumption that each input features are considered independent of the others.

## 8. Comparison with Literatures

The CBR, UCS and R-value are vital mechanical properties of subgrade soil that researchers have proposed several ML techniques to predict. It is noteworthy that several of these studies have applied either the whole or part of the 121 dataset applied in this study. Therefore, a need for comparison of our results with those reported in literatures. It should be noted that establishing a fair comparison of performance with literatures can be difficult due to differences in sample sizes, training-test split ratios (e.g., some may use 80:20 ratio while others may use 90:10 or 70:30), statistical evaluation metrics and experimental test setup. Regardless of these differences, these concerns have been taken into account with the extensive nature of the experiments done in this study. Table 4 shows the performance of results in literatures in detail giving the data size, experimental design, and algorithm, in terms of $R^2$, RMSE, MAE and MAPE.

Table 4 shows the comparison of the results obtained in this study with the best in literatures. An observation of the table indicate that several ML techniques have been applied to predict the CBR, UCS and R of subgrade soils. Considering the performance of these studies, the best performance for predicting CBR found in the reported previous studies is found in [23] where the applied ANN yielded $R^2$= 0.9994, RMSE = 1.1900, and MAE = 0.1649. The best performance from the three ML techniques in this study is XGB1 with $R^2$= 0.9999, RMSE = 0.1119, and MAE = 0.0926. The XGB1 result outperformed the best result from previous studies by improvements of 0.05%, 90.5%, and 43.8% in the $R^2$, RMSE, and MAE values respectively. Furthermore, the SVR1 and CBT1 models in this study also yielded better results than Onyelowe [23].

Regarding UCS prediction, the ANN model from [14] reported the best results in terms of with $R^2$ (0.9992) and RMSE (0.5570) whereas the Fuzzy Logic model from the same study produced the best MAE (0.3145). Compared to the best result in this study, while the XGB2 model yielded a better result than the ANN model in terms of $R^2$ (0.9995) and MAE (0.6081), the MAE remains inferior to the that of the Fuzzy logic model from the same study. It should also be noted that both CBT2 and SVR2 models in this study produced better $R^2$ compared to [14].

For R-value prediction, the $R^2$ = 0.9999, RMSE = 0.0556, and MAE = 0.0444 produced by the XGB3 model is better than the best methods from previous studies except in terms of MAE, where the MAE = 0.0380 reported in [23] is better. SVR3 and CBT3 are equally better than the best from the previous studies in term of $R^2$ and RMSE.

Overall, the XGB-based models performance is quite impressive in predicting the mechanical properties of subgrade soils, alluding to the robustness across different prediction tasks [31]. Nevertheless, the results given in this work is confined to these three mechanical properties (CBR, UCS and R) of subgrade soils.

Table 4. Results comparison with literatures

| Researcher | Algorithm | Dataset size | $R^2$ | RMSE | MAE | MAPE |
|---|---|---|---|---|---|---|
| **CBR** | | | | | | |
| [23] | ANN | 121 | 0.9994 | 1.1900 | 0.1649 | - |
| [14] | ANN | 121 | 0.9987 | 0.4346 | 0.2987 | - |
|  | Fuzzy Logic |  | 0.9921 | 0.5561 | 0.3213 | - |
| **This Study** | **XGB1** | 121 | **0.9999** | **0.1119** | **0.0926** | **0.0048** |
|  | **CBT1** |  | 0.9994 | 0.2868 | 0.2135 | 0.0101 |
|  | **SVR1** |  | 0.9997 | 0.2191 | 0.1605 | 0.0080 |
| **UCS** | | | | | | |
| [23] | ANN | 121 | 0.9350 | 1.1900 | 1.2700 | - |
| [14] | ANN | 121 | 0.9992 | **0.5570** | 1.3230 | - |
|  | Fuzzy Logic |  | 0.9981 | 0.8152 | **0.3145** | - |
| **This Study** | **XGB2** | 121 | **0.9995** | 0.7343 | 0.6081 | 0.0039 |
|  | **CBT2** |  | 0.9994 | 0.7985 | 0.6082 | 0.0035 |
|  | **SVR2** |  | 0.9988 | 1.1039 | 0.5466 | **0.0031** |
| **R** | | | | | | |
| [23] | ANN | 121 | 0.9900 | 1.1900 | **0.0380** | - |
| [14] | ANN | 121 | 0.9970 | 0.2545 | 0.2033 | - |
|  | Fuzzy Logic |  | 0.9810 | 0.6251 | 0.4852 | - |
| **This Study** | **XGB3** | 121 | **0.9999** | **0.0556** | 0.0444 | **0.0023** |
|  | **CBT3** |  | 0.9996 | 0.0995 | 0.0764 | 0.0043 |
|  | **SVR3** |  | 0.9997 | 0.0794 | 0.0672 | 0.0036 |

## 9. Conclusions

This study presents CatBoost, XGBoost and support vector regression-based models for the prediction of the CBR, UCS, and R of subgrade hydrated-lime activated rice husk ash-treated soil. Although, the model estimations of the mechanical properties were mostly in good agreement with the experimental values, evaluation results based on four statistical measures showed the superiority of the XGBoost-based models in the estimation of these properties compared to the CatBoost and support vector regression-based models, given the former's mostly lower errors and higher coefficient of determination. A sensitivity analysis of the impact of each input feature on the prediction of each model is also presented using partial dependence plots, with all models agreeing that HARSH proportion influences the model prediction. The superiority of the proposed models is shown by the comparison made with findings from previous studies.of results with findings from previous studies also shows the superiority of the models, particularly in CBR and R-value estimations. The study presents efficient ML techniques to the growing toolkits of site engineers for the estimation of design properties of subgrade soils.

**Author Contributions**
**Ismail B. Mustapha:** Investigation, methodology, writing-original draft. **Muyideen Abdulkareem**: Supervision, conceptualization, writing-original and final drafts. **Abideen Ganiyu:** Data curation, writing-original draft, writing- review & editing. **Hatem Nabus:** Conceptualization, data curation, investigation, methodology, software. **Jin Chai Lee:** Data curation, formal analysis, software.

**Conflict of interest statement**
The authors declare no conflicts of interest.

**Data availability statement**
Data are available upon reasonable request from the corresponding author.